%% file: ewrl_2026.tex
\documentclass{article}

\usepackage[final]{ewrl_2026}

\usepackage[utf8]{inputenc} 
\usepackage[T1]{fontenc}    
\usepackage{hyperref}       
\usepackage{url}            
\usepackage{booktabs}       
\usepackage{amsfonts} 
\usepackage{amssymb}
\usepackage{amsthm}
\usepackage{amsmath}
\usepackage{nicefrac}       
\usepackage{microtype}      
\usepackage{xcolor}         
\usepackage{graphicx}
\usepackage{subcaption}

\newtheorem{theorem}{Theorem}[section]
\newtheorem{lemma}[theorem]{Lemma}

\include{my_commands}

\title{Exploration and Online Transfer with \\
Behavioral Foundation Models}

\author{%
  Louis Bagot \\
  Université Lyon 1, CNRS, INSA Lyon\\
  LIRIS - UMR5205, 69622 Villeurbanne, France \\
  \texttt{louis.bagot@univ-lyon1.fr} \\
  \AND
  Mathieu Lefort \\
  Université Rennes, Inria, CNRS\\
  IRISA - UMR 6074 \\ 
  F-35000 Rennes, France\\
  \And
  Laetitia Matignon \\
  Université Lyon 1, CNRS, INSA Lyon\\
  LIRIS - UMR5205 \\
  69622 Villeurbanne, France \\
}

\begin{document}

\maketitle

\begin{abstract}
Zero-shot Transfer in Reinforcement Learning (RL) aims to train an agent that can generate optimal policies for any reward function, without additional learning at transfer time, while training only on reward-free trajectories. 
For their generality over tasks, such models are sometimes called ``Behavioral Foundation Models'' (BFMs). 
While they have shown strong performances and improvements in recent years, the current framework and algorithms still assume that, during the transfer phase, the agent is informed \textit{offline} about the reward (the task to solve) through a dataset of state-reward pairs, which it uses to pick the best policy to deploy. 
However, in practice if the reward is a black-box (e.g. direct user feedback), it is not possible to generate such a dataset: it is necessary to observe the reward through interactions with the environment.
In other words, the current framework of offline transfer is not aligned with the traditional RL setting of online learning through trial-and-error, which requires exploration in order to find rewards.
This paper proposes to tackle this new \textit{online} transfer in zero-shot RL, with the key insight that the BFM itself can be used to generate exploration policies.
We show that it is possible to frame this online learning problem in terms of a bandit-like exploration-exploitation problem. 
More precisely, at each step the bandit algorithm recommends a policy, the BFM executes it in the environment, which yields a reward and a new state; we repeat the process until we converge to the optimal policy.
In the popular context of linear reward approximation, we derive a formulation inspired by Upper Confidence Bound and show that exploration can be achieved through the minimization of the eigenvalues of an uncertainty matrix. We evaluate qualitatively and quantitatively our framework on a simple environment to validate the concept of our method.
\end{abstract}

\section{Introduction} \label{sec:intro}
In Reinforcement Learning (RL, \citet{sutton2018}), the \textit{agent} learns to interact with its \textit{environment} and gather \textit{rewards} through trial-and-error. 
Zero-shot Transfer in RL \citep{touati2023does} is an extension of RL focused around \textit{task transfer}: quickly finding the optimal policy for any new reward function (any new \textit{task}).
Zero-shot RL is formalized as follows: in a \textit{pre-training} phase, the agent is trained on interactions with the environment but \textit{without rewards}. 
Then, during the \textit{transfer} phase, the agent must maximize a reward function without additional training.
Zero-shot RL therefore requires the learning of general \textit{knowledge} about the reward-free environment, which may be useful to as many downstream tasks as possible.
This field has seen a lot of attention in the past few years \citep{park2024hilp,frans2024fre,bagot2025successorclusters,agarwal2025proto}, coining the term \textit{Behavioral Foundation Models} (BFM, e.g. \citet{tirinzoni2025zero,li2025bfm}) due to the large diversity of behaviors they can generate. 
In this paper we focus on the transfer phase, so we assume access to a pre-trained BFM. 

In current zero-shot RL practice, during the transfer phase, the agent is informed about the task to solve through a \textit{transfer dataset} of state-reward pairs. We call this type of transfer \textit{offline}, because the agent did not need to interact with the environment to understand the task.
However, if the reward is expensive to compute, black-box or can only be obtained through interactions with environment (e.g., direct human feedback or electricity prices), it is not possible to generate the transfer dataset ahead of time. 
In this case, the agent needs to explore and understand the task \textit{during} the transfer phase.
Such \textit{online} transfer is closer to the standard RL framework, where the agent needs to learn by trial-and-error through its interactions with the environment, and needs to \textit{explore} to find rewards.
How does zero-shot RL deal with such situations then? From \citet{touati2021fb}: ``\textit{Thus, if the reward is black-box as in standard RL algorithms, then the exploration policy has to be run again for some time}''. 
The idea is to use a generic exploration policy to gather the transfer dataset before running the BFM as in offline transfer. 
However, in general we only have access to the random policy, which will not explore efficiently: it may take a very long time to observe the relevant parts of the environment. 
This would also require picking a satisfactory size for the transfer dataset, which may depend on the quality of exploration and how informative it is about the task to solve.
Importantly, a generic exploration policy would explore the space irrespective of what the BFM can actually do: for example if our BFM focuses on position but not speed, we do not need to explore different speeds.

\begin{figure}[!t]
    \centering
    \includegraphics[width=.8\textwidth]{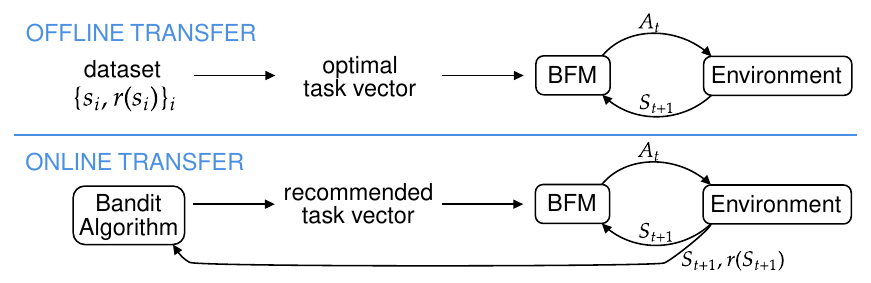}
    \caption{Difference between \textit{offline} transfer, as standard in zero-shot RL literature, and the \textit{online} transfer we propose and study. In offline transfer, a dataset of state-reward pairs enables the direct computation of the optimal task vector which the Behavior Foundation Model (BFM) uses as conditioning to execute the optimal policy. However, in practice, we often cannot generate such a dataset, so interactions with the environment are necessary: this is online transfer. We propose to view it as a Bandit-like framework where an Algorithm must \textit{recommend} a task vector to the BFM, and use the reward feedback from the environment to converge to the optimal task vector.
    }
    \label{fig:framework}
\end{figure}

We aim to tackle online learning more efficiently. Specifically, we ask:
\textbf{can we leverage the flexibility of the Behavioral Foundation Model itself to make it generate efficient exploration policies?}
In practice, BFMs generally use a \textit{task vector} as conditioning to perform the task. Offline transfer is efficient because we can directly compute the \textit{optimal task vector} from the transfer dataset.
To perform online transfer, we propose to judiciously query the BFM on different task vectors to generate policies which efficiently explore the state space. 
This disposes of the need for a generic exploration policy and, as we will show, through our method the BFM is able to explore to rapidly gain information \textit{about the optimal task vector}, so exploration is guided around what the BFM can do.
To achieve this, we propose to re-frame this problem as a Bandit (or recommendation) problem \citep{lattimore2020bandit} where, at each time-step, an algorithm recommends a task vector, which is executed by the BFM in the environment to obtain a reward (Figure \ref{fig:framework}). This sequence of recommendations should converge quickly to the optimal task vector. 
In the context of BFMs based on linear reward approximation, one of the most popular approaches for BFMs, we propose an exploration-exploitation algorithm inspired by Upper Confidence Bound (UCB) to balance exploration and exploitation.

In Section \ref{sec:background} we set the theoretical grounds for our work. Our contributions are the following: 
\textit{(i)} we introduce a new problem (online transfer from zero-shot policies, aka BFMs) and propose to tackle it as a bandit problem (Sections \ref{subsec:problem_statement} and \ref{subsec:bandit}). 
\textit{(ii)} We introduce a novel and practical optimization algorithm (USF-UCB) for picking a sequence of conditioning vectors in the context of online transfer under linear reward approximation (Sections \ref{subsec:linear_approximation} and \ref{subsec:approach}), for which we derive theoretical results. 
\textit{(iii)} We test our algorithm on a simple environment to provide a proof-of-concept and show that it does produce behaviors that balance exploration and exploitation (Section \ref{sec:experiments}).
To round up the paper, we talk about related work in section \ref{subsec:related_work}, and conclude in Section \ref{sec:discussion}.


\section{Background} \label{sec:background}

Our paper studies online transfer in zero-shot transfer in Reinforcement Learning. We start by introducing Reinforcement Learning and its zero-shot version, which gives the theoretical basis for the Behavioral Foundation Models with which we want to perform transfer. We then introduce Universal Successor Features, a popular approach for zero-shot RL which we use to derive our algorithm. Since our work is inspired by Linear Bandits, we give an introduction in Appendix \ref{app_subsec:linear_bandits}.

\textbf{Reinforcement Learning (RL)} \citep{sutton2018} is formalized with a Markov Decision Process (MDP) $\gM = \left(\sS, \sA, r, p, \gamma, \mu_0 \right)$, where $\sS$ is the set of states, $\sA$ is the set of actions, $r$ is the reward function, $p$ is the \textit{dynamics} or \textit{transition} function $p\left(s^\prime \mid s,a\right)$ as a probability distribution over next states, $\gamma \in [0,1[$ is a discount factor and $\mu_0$ is the distribution over initial states. The agent is modeled as a \textit{policy}, a distribution from states to actions $\pi \left(a \mid s\right)$. This completes the setup: a \textit{trajectory} is the sequence of random variables generated from the interactions of $\mu_0$, $\pi$ and $p$: $\tau = S_0, A_0, R_1, S_1, A_1, R_2, S_2,\dots$
We can then define the \textit{value function}, the expected sum of discounted rewards when following a policy $\pi$: $v_\pi \left(s\right) \doteq \E_\pi \left[ \sum _{k=0}^\infty \gamma ^k r(S_{t+k+1}) \mid S_t=s \right]$. The objective of RL is to find the policy with the highest value function. 

\textbf{Zero-shot RL} is a RL framework with additional constraints. We follow the most popular framework as formalized by \citet{touati2023does}: ``\textit{the goal of zero-shot RL is to compute a compact representation $\mathcal{E}$ of the environment by observing samples of reward-free transitions $(s_t, a_t, s_{t+1})$ in this environment. Once a reward function is specified later, the agent must use $\mathcal{E}$ to immediately produce a good policy, via only elementary computations without any further planning or learning.}'' 
In order to tackle this challenge, most zero-shot approaches rely on a \textit{encoding-decoding procedure} of the reward function, $r \longleftrightarrow \z_r \in \sR^d$, where the \textit{task vector} $\z_r$ encodes the task (either the reward $r$ itself or an optimal policy). This encoding is used to train a task-conditioned policy $\pi \left(a\mid s,\z_r\right)$ which must provide an optimal policy for reward function $r$ (see Appendix \ref{app:encoding_decoding} for a more precise formalization).
Such $\pi$ are sometimes called \textit{Behavioral Foundation Models} \citep{tirinzoni2025zero,li2025bfm}.

\textbf{Universal Successor Features} (USFs, \citet{barreto2017successor,barreto2018transfer}) 
are one of the most popular frameworks for zero-shot RL. SFs propose to linearly approximate the reward function:
assume access to a linear basis over rewards, a set of $d$ \textit{features} $\boldphi : \sS \rightarrow \sR^d$. We can treat them as pseudo-rewards and consider their value functions, the \textit{Successor Features}: $\boldpsi_\pi (s) \doteq \E_\pi \left[ \sum _{k=0} ^\infty \gamma ^k \boldphi \left(S_{t+k+1}\right) \mid S_t=s \right]$. 
Critically, if we consider a reward function $r(s)= \boldphi(s)\cdot \z_r$ for some vector $\z_r \in \sZ \subset \sR^d$, then the linear property carries over to the action-value function:  $v_\pi (s) = \boldpsi_\pi \left(s\right) \cdot \z_r$, granting easy policy evaluation.
Note that now, given $\boldphi$, $\z_r$ fully encodes the reward function: we can consider the policy optimal for any $\z$, $\pi \left(a\mid s,\z\right)$ and its \textit{Universal Successor Features}, $\boldpsi (\cdot;\z) \doteq \boldpsi_{\pi(\cdot; \z)}$. Now $\boldpsi (\cdot;\z)$ encodes the future features that will be visited under the optimal policy for \textit{any} choice of $\z$ \citep{borsa2018universal}. Several zero-shot RL methods use this framework and are distinguishable by their choice of features $\boldphi$. In this paper we pick two to experiment with, which we detail in Appendix \ref{app_sec:lap_clusters} and Figure \ref{fig:features_usfa}.

\section{Framework and Method} \label{sec:method}

\subsection{Problem Statement} \label{subsec:problem_statement}
Assuming access to a pre-trained Behavior Foundation Model (BFM), our objective is to perform online transfer. This means we want to find the optimal policy associated with transfer reward $r$ through online interactions with the environment, i.e., under a black-box reward and without preemptive access to a transfer dataset of state-reward pairs (Figure \ref{fig:framework}).
We write the BFM as $\pi (a \mid s,\z)$, the optimal policy for some task vector $\z \in \sZ \subset \sR^d$. We call $\z_r$ the optimal task vector, in other words, $\pi(\cdot,\z_r)$ is optimal for the transfer reward $r$. We are looking for an algorithm to quickly find $\z_r$ from online interactions. 
Critically, in zero-shot fashion, we are not interested in fine-tuning or re-training the BFM $\pi$: we want to exploit the knowledge and behaviors it contains without re-training it.

For our algorithm we specifically assume access to all components of the USF framework as detailed in the previous Section, \ref{sec:background}. To re-iterate: the features $\boldphi$; the policy (BFM) $\pi (a \mid s,\z)$ optimal for the reward functions expressed as the linear combination $r(s) = \boldphi(s) \cdot \z$,  for any $\z \in \sZ$; the USFs, $\boldpsi (\cdot;\z)$, with property $\boldpsi (\cdot;\z) \cdot \z = v_{\pi(\cdot, \z)}$. 
Following standard practice, we assume that $\sZ$ is the sphere of radius $\sqrt{d}$: the task vectors are normalized before being passed to the BFM.

\subsection{Bandit formalization and objective} \label{subsec:bandit}

\textbf{Bandit-MDP Interactions.}
Starting from $S_0 \sim \mu_0 (\cdot)$, at each time-step $t$, from state $S_t$, the algorithm must choose a conditioning vector $Z_t$, which is passed to the BFM to generate $A_t \sim \pi \left(\cdot \mid S_t, Z_t\right)$. 
The environment then responds with the next state $S_{t+1} \sim p\left(\cdot \mid S_t, A_t\right)$ and noisy reward $R_{t+1} = \boldphi(S_{t+1})\cdot \z_r + \eta _{t+1}$ with $\eta_t \sim \N (\boldsymbol{0},\sigma^2I) $. 
We therefore have the \textit{trajectory} $\tau_t = S_0, Z_0, A_0, S_1, R_1, Z_1, A_1, \dots S_t$ (assuming no termination).

\textbf{Objectives for online transfer with arbitrary BFMs.}
We are looking for an algorithm $\Lambda (Z_t \mid \tau_t)$ to generate the next conditioning vector given the trajectory so far. 
As a reminder, we are interested in finding $\z_r$, with which $\pi$ is optimal. Instead of focusing on instantaneous rewards, we propose an objective which asks, at each time-step, ``how good would it be to execute \textit{policy} $\pi (\cdot ;Z_t)$?'', formalized as:
\begin{align}
    \min_{\Lambda} ~ \text{Reg}_{T} \doteq \sum_{t=1}^{T-1}v_{\pi\left(\cdot,\z_{r}\right)}\left(S_{t}\right)-v_{\pi\left(\cdot,Z_{t}\right)}\left(S_{t}\right)
\end{align}
We borrow from bandit theory the notion of \textit{regret} $\text{Reg}_{T}$ as a distance from theoretical perfection: $v_{\pi\left(\cdot,\z_{r}\right)}$ is the value function for the optimal policy, and each time-step spent playing anything but $\z_r$ is sub-optimal.
From this objective, a direct approach would be to entirely unroll the policy $\pi(\cdot, Z_t)$ to evaluate its value function, and use this feedback to pick the next recommendation. However, this would require waiting until the end of the episode or some large time $T$ between each decision. Instead, we would like to learn from each time-step in the environment.
The linear decomposition of Successor Features allows us to achieve this, by updating our estimate of $\z_r$ with each instantaneous reward observed. We therefore turn to this framework:

\textbf{Objectives for online transfer with USFs.}
 Through the linear decomposition of the value function, we get the following final objective:
\begin{align}
	\min_{\Lambda} ~ \text{Reg}_{T} = \sum_{t=1}^{T-1}\boldpsi\left(S_{t},\z_{r}\right)\cdot \z_{r}-\boldpsi\left(S_{t},Z_{t}\right)\cdot \z_{r}. \label{eq:regret}
\end{align}
Equation \eqref{eq:regret} is reminiscent of Linear Bandits (see Appendix \ref{app_subsec:linear_bandits} for a formal introduction), and the corresponding Chapters in \citet{lattimore2020bandit} have inspired the rest of the method. In Appendix \ref{app_sec:bandit_comparison} we take a step back to compare our framing and objective to that of Linear Bandits.

\subsection{Linear Estimation and Uncertainty Matrix} \label{subsec:linear_approximation}
\textbf{Linear Estimator.}
We are looking for the sequence of $Z_t \in \sZ$ which converges quickly to $\z_r$. 
Since we use the USF framework where $R_t=\boldphi (S_t)\cdot \z_r + \eta _t$, we have a clear estimate for $\z_r$ as the regularized linear least squares estimator: 
$ \hat{Z}_{t}=\arg\min_{\z}\sum_{k=1}^{t}\left(R_{k}-\boldphi\left(S_{k}\right)\cdot \z\right)^2+\lambda\left\Vert \z\right\Vert _{2}^{2} $, 
with penalty factor $\lambda \geqslant 0$ for regularization. $\hat{Z}_t$ has a well-known analytical solution:
\begin{align}
	\hat{Z}_{t}&=V_{t}^{-1}\sum_{k=1}^{t}\boldphi\left(S_{k}\right)R_{k} & \text{with }V_{t}&=\lambda I+\sum_{k=1}^{t}\boldphi\left(S_{k}\right)\boldphi\left(S_{k}\right)^{\top}. 
\end{align}
From our estimate $\hat{Z}_t$ we can build a confidence bound, which contains $\z_r$ with high probability: from bandit theory, it will generally have the shape 
\begin{align}
	\varepsilon_{t}=\left\{ \z\in\sZ\mid\left\Vert \z-\hat{Z}_{t}\right\Vert _{V_{t}}^{2}\leqslant\beta_{t}\right\}. \label{eq:ellipsis}
\end{align}
In other words, $\varepsilon_{t}$ is an ellipsis centered around $\hat{Z}_{t}$ and whose directions are controlled by the eigenvectors and eigenvalues of $V_t ^{-1}$. The diameter is controlled by $\beta_{t}$, for which naive forms exist \citep{lattimore2020bandit}; we leave finding tighter forms to future work.

\textbf{General Approach to Exploration.}
For now we take a moment to interpret these definitions in our context. In order to find the best $\z_r$, we need to explore the different directions of the feature space encoded by $\boldphi (\sS)$. As we gather more information about the domain and reward, the directions we visit are stored in $V_t$, which we can interpret as a sort of \textit{certainty matrix} (sometimes called the \textit{information}, \textit{Gram} or \textit{design} matrix). In the case of cluster occupancy features (see Appendix \ref{app_sec:lap_clusters}), this has a direct interpretation: since $\phi_i(S_t)=\boldsymbol{1}_{S_t \in C_i}$, the matrix $V_t$ is a diagonal matrix where each diagonal component is the number of times that cluster $C_i$ was visited in the trajectory so far. By construction, ellipsis $\varepsilon_t$ is controlled in each direction by the eigenvalues and eigenvectors of $V_t ^{-1}$, the \textit{uncertainty matrix}. 
For cluster features, this leads to an inverse count, which also directly correspond to its eigenvalues: the more we visit a cluster, the more we reduce our uncertainty about it. 
Our general approach to exploration will therefore consist in \textbf{reducing the uncertainty}, or more formally, \textbf{minimizing the eigenvalues of }$V_t ^{-1}$. This will shrink the ellipsis $\varepsilon_t$ and converge to $\z_r$. 

\subsection{Approach to Exploration with USF-UCB} \label{subsec:approach}
\textbf{Proposed Algorithm.}
At any time-step $t$ the greedy option is to choose $Z_t = \hat{Z}_{t}$, our best estimate of $\z_r$. However, our estimator is not perfect and improving it requires exploration, in particular in the directions of the space we have least explored, as encoded by the eigenvalues and eigenvectors of $V_t ^{-1}$. Finding a $\z$ which is beneficial for exploration comes down to asking ``if we were to roll out policy $\pi (\cdot, \z)$, which $\z$ would lead to the most uncertainty reduction?''. 
For this we use a common object in bandits, the \textit{elliptical norm} $\left\Vert \boldsymbol{x} \right\Vert_{V_t ^{-1}} ^2 \doteq \boldsymbol{x}^\top V_t ^{-1}\boldsymbol{x}$ (sometimes called \textit{energy} or \textit{Mahalanobis} norm). For two vectors $\boldsymbol{x_1},\boldsymbol{x_2}$ of same $L^2$ norm, the one with higher elliptical norm under $V_t ^{-1}$ is the one which aligns most with the eigenvectors with highest eigenvalues of $V_t ^{-1}$. In other words, this norm allows us to measure how uncertain a direction is. 
We employ an approach similar to Upper Confidence Bound (UCB), where we pick the most uncertain option; this leads to the following algorithm, which we call USF-UCB:
\begin{align}
	Z_t &= \underset{\z \in \sZ}{\arg \max} \, \boldpsi\left(S_{t}; \z\right)\cdot \hat{Z}_{t} + \sqrt{\beta_t} \left\Vert \boldpsi \left(S_t, \z\right) \right\Vert_{V_{t} ^{-1}}. \label{eq:USF-UCB}
\end{align}
We want to find the vector which both yields a high return under our linear estimator $\hat{Z}_{t}$ (first term), and explores the uncertain parts of the space (second term). 
Importantly, \textbf{the norm $\left\Vert \boldpsi \left(S_t, \z\right) \right\Vert_{V_{t} ^{-1}}$ measures how informative it would be to unroll the \textit{policy} $\pi (\cdot, \z)$}, as expressed by the uncertainty of the features it visits.
We find the following bound on regret for this algorithm:
\begin{lemma}
    The regret of USF-UCB as Equation \eqref{eq:USF-UCB} is bounded as follows:
    \begin{align}
        \text{Reg}_T \leqslant 2\sum _{t=1} ^{T-1} \sqrt{\beta _t} \left\Vert \boldpsi \left(S_t, Z_t\right) \right\Vert_{V_{t} ^{-1}}.
        \label{eq:regret_bound}
    \end{align}
\end{lemma}
See Appendix \ref{app_sec:lemma_proof} for the derivation. In linear bandits regrets involving the elliptical norm are bounded above through the \textit{elliptical potential lemma} \citep{abbasi2011improved,carpentier2020elliptical} by relating the sum of norms to the determinants of $V_T$ and $V_0$. We explore this direction in Appendix \ref{app:elliptical_potential_lemma} but leave finding a proper bound to future work.
It may seem counter-intuitive to maximize in Equation \eqref{eq:USF-UCB} the quantity we want to minimize in Equation \eqref{eq:regret_bound}, but this is a common approach in bandits: we maximize the instantaneous uncertainty so that the sum of future uncertainties is smaller. In Appendix \ref{app:more_ellipsis} we discuss other angles through which the maximization of the elliptical norm comes about.

\textbf{Optimization.}
Equation \ref{eq:USF-UCB} requires optimization to find the argmax. Since in practice $\boldpsi$ is a trained neural network, this is possible through gradient descent (freezing the network weights), though our experiments find that it leads to unstable results. We suspect this is mainly due to over-approximation errors, and for our simple experiments we find instead that sampling several random $\z \sim \sZ$ and picking the ones with highest elliptical norms yields satisfactory results. We will study more advanced optimization methods with future work and sturdier evaluation metrics and environments.

\input{fig_pretrain}

\section{Experiments} \label{sec:experiments}
\subsection{Domain and pre-training setup}
We test our method in a simple $9\times 9$ gridworld with actions \texttt{up}, \texttt{down}, \texttt{left}, \texttt{right} and blocking edges. 
While this environment is generally trivial to solve for a given reward function, generalizing over all possible task vectors $\z$ is not simple for a neural network. 
We implement two features choices for the Successor Feature basis $\boldphi : \sS \rightarrow \sR^d$: cluster occupancy features $\phi_i(s)=\boldsymbol{1}_{s \in C_i}$ and Laplacian eigenfunctions $\phi_i(s)=f_i(s)$ which we detail in Appendix \ref{app_sec:lap_clusters}. We pick $d=9$ as a compromise of expressivity and ease of visualization. 
We detail the training of the \textit{USF Approximator} (USFA) $\boldpsi _{\boldtheta} \left(S_t,\z \right)$ in Appendix \ref{app_sec:usfa}. 
We visualize the features and some of the trajectories produced by the trained USFA in Figure \ref{fig:features_usfa}. We can clearly see that the task-conditioned policy is not perfect, but achieves satisfactory performance on most tasks. 

\subsection{Pure Exploration}
The crux of the algorithm is in finding a sequence of $Z_t$ to explore the domain, so we start by focusing on this and set $\z_r=\boldsymbol{0} = \hat{Z}_t$ for this sub-section -- in other words, the agent is only motivated to explore.
We optimize Equation \eqref{eq:USF-UCB} by argmaxing over $10000$ random samples in a single batch. 

\textbf{Visualization of trajectories.} We start by visualizing exploration trajectories for both feature choices in Figure \ref{fig:explo} to qualitatively evaluate the method. We find that the exploration is efficient but not perfect: it does not span every single state. This is expected: for cluster occupancy features, it is sufficient to walk anywhere in the cluster for the feature to be active, so the agent has no incentive to reach the far-away corners. In other words, the exploration is efficient \textit{from the perspective of visiting clusters}, and in general \textbf{our algorithm explores efficiently with respect to its own features}. While this is a limitation, this can also be desirable by restricting our exploration to only the parts that our features cover (and therefore our BFM can differentiate for control). 

\textbf{Eigenvalue analysis.}
We study the algorithm from the perspective of the objective it tries to optimize: the minimization of the uncertainty, which we measure with the determinant of $V_t^{-1}$. In order to achieve this, we compare the method to an ``exhaustive'' explorer, which visits all states (details and visualization in Appendix \ref{app:exhaustive}), as well as a random explorer. We visualize the results for both feature choices in Figure \ref{fig:determinants}, measuring the determinant first with the own method's features and next with state visits (uncertainty matrix of $d=\left\vert \sS \right\vert$). We find again that each method performs well for its own class of features: the USFAs with our proposed method explore their features as fast or faster than the exhaustive agent, but do not manage to reach all states. This aligns with our qualitative intuition from the previous paragraph. 

\subsection{Online transfer}
In this section the agent must maximize a reward function by discovering its values in the state space. The instantaneous reward comes as $R_{t}=\boldphi (S_{t})\cdot \z + \eta_t$ with Gaussian $\eta_t$ at $\sigma=0.3$. We test our algorithm as Equation \eqref{eq:USF-UCB}. In Linear Bandits, the ellipsis radius $\beta_t$ is a well-studied object with analytical values. Due to the reliance on the non-linear USF, this analysis is more complicated in our case and we leave this direction for future work. After some fine-tuning, we choose $\beta_{t<50}=1$ to encourage early exploration, and $\beta_{t>150}=0$ to allow convergence, with linear annealing in-between.

\textbf{Qualitative evaluation.} We hand-pick $\z_r$ and visualize online transfer trajectories in Figure \ref{fig:samples_exploit}. Note that we now only display $\hat{Z}_t$ in the background, visualizing the agent's best linear estimate converge over time. Qualitatively, the estimate seems to improve and the trajectory ends in the highest-rewarding states. We now test this more systematically.

\textbf{Quantitative analysis.} We evaluate the method through the estimate quality and reward obtained. We sample $10$ random $\z$ and run online transfer with USF-UCB to compute the L2 distance between $\z_r$ and $\hat{Z}_t$, as well as the instantaneous rewards. We plot the results in Figure \ref{fig:exploit}. We can see that the estimate reaches very close to the true value despite the added noise, and the trajectory ends up maximizing reward. 
Note that the agent is able to execute the optimal policy after observing around $100-150$ samples. In comparison, a natural approach for offline transfer in this case would be to fix a dataset containing all $81$ states; from that perspective the online approach (at least in this simple environment) is not prohibitively sample-expensive. 

\include{fig_expes}

\section{Related work} \label{subsec:related_work}

Our framework is tightly related to \textbf{Hierarchical RL} \citep{hutsebaut2022hierarchical} which proposes to break the RL problem down into sub-problem using sub-policies; specifically the \textit{options} framework \citep{sutton1999between} where an option is a temporally extended action: a ``meta''-agent has access to an option set $\sO$, where each option $o$ calls another policy (with initiation and termination conditions). In our case we treated $\sZ$ as the option set on which our agent acts, abstracting out the BFM policy underneath, similarly as \citet{barreto2019option}. Instead of training a meta-policy $m(\z\mid s)$ with different options depending on the current state, we assume that there is a single optimal $\z_r$, which avoids complicated value function estimation. Agent57 \citep{badia2020agent57} proposed to optimize the weight $\beta$ between intrinsic and extrinsic reward with a bandit algorithm, which has similarities with our setup with $d=1$, though this was used during training instead of at transfer-time. 

\textbf{Continual RL} \citep{khetarpal2022towards} 
is a framework of RL where all elements of the MDP may change over time and the agent needs to continuously adapt. Zero-shot RL is therefore a specific case of Continual RL where only the reward function changes with time in a specific way ($0$ during pre-training, constant in time during transfer). This viewpoint opens questions on the ability for zero-shot RL to \textit{(i)} develop skills continuously instead of through a fixed dataset, and \textit{(ii)} continuously adapt to new reward functions. Our paper contributes to the second question.

\textbf{Intrinsic motivation.} Generating bonuses for exploration is a common approach in RL, sometimes called intrinsic motivation. While bonuses involving the uncertainty matrix have been proposed in RL \citep{jin2020provably,bai2021principled}, these methods assume linearity in both the reward and dynamics of the environment, a strong assumption which permeates the analysis. The USF framework allows us to work with general unknown dynamics and maximize exploration \textit{over a family of candidate policies} instead of focusing on instantaneous uncertainty. When using successor measures for features, our approach generalizes inverse state or region counts, which is a very popular approach to intrinsic motivation in RL \citep{strehl08mbieeb,bellemare2016unifying}

\section{Conclusion} \label{sec:discussion}
\subsection{Summary}
We have presented a novel framework and approach for addressing online transfer with Behavioral Foundation Models. In particular, we have proposed to frame the scenario similarly to a linear bandit problem, where an algorithm must recommend policies at each timestep in order to explore the state space, discover rewards and converge to the optimal policy. We have provided the essential theoretical building blocks and a well-justified algorithm to minimize regret through uncertainty minimization, inspired by Upper Confidence Bound. We have evaluated our method in a simple domain and showed that we can efficiently explore and exploit by manipulating a Behavioral Foundation Model, enabling online transfer and taking a step towards the continual learning of such models. 

\subsection{Limitations and future work}
The clearest step forward is to show similar results on much more complicated environments, with stronger baselines, over a wider range of feature choices (e.g. \citet{touati2023does,park2024hilp}), to demonstrate that the method scales, and this will be our next direction. As a second direction, our work is currently extremely dependent on the feature quality and performance of the Behavioral Foundation Model. If the reward we're trying to solve is not encoded by the features, or the BFM did not properly learn an optimal policy, we cannot proceed. Our USF-UCB algorithm also heavily relies on the USFA to provide a good approximation of future feature visits, while in practice these approximations may be unreliable. It could be necessary to adapt the algorithm to make it more resilient against poor estimations, for example through averaging over regions or using ensemble methods. Finally, more theoretical work can be done on the framework we propose, mainly on the form of the regret and confidence bounds, and potentially on alternative algorithms. Following linear bandit theory, future work could tighten the regret bound through the elliptical potential lemma \citep{carpentier2020elliptical}.

\newpage
\bibliographystyle{apalike}
\bibliography{ref}

\appendix

\section{Theoretical framework} \label{app_sec:theoretical_framework}

\subsection{Encoding-decoding procedure for zero-shot RL} \label{app:encoding_decoding}

In Section \ref{sec:background} we defined the zero-shot RL problem where an agent must accumulate knowledge about the reward-free environment in a pre-training phase, and use this knowledge to instantly provide an optimal policy during the transfer phase, without additional training. 

We now clarify the ``encoding-decoding procedure'' used by most zero-shot methods.
\begin{itemize}
    \item In RL, the objective is specified in the reward function $r$, for which the agent must find an optimal policy. The objective for zero-shot RL can be seen as training a \textit{task-conditioned policy} $\pitheta \left(a \mid s,r\right)$ to be optimal for $r$. 

    \item Of course, in general $r$ cannot be handled directly, so the idea is to encode the function into a vector $r \longrightarrow \z_r \in \sZ \subset \sR^d$ and train policy $\pitheta \left(a \mid s, \z_r\right)$. 

    \item In practice, since we still cannot handle $r$ directly the encoding happens through labeled state-reward pairs in the transfer dataset $\mathcal{D}_{\textbf{transfer}} \doteq \{s_i, r(s_i)\}_{i=1}^ D$. In other words, the encoding function is $\epsilon(\mathcal{D}_{\textbf{transfer}}) = \z_r$. For transfer, we simply need to execute $\pitheta \left(\cdot, \z_r\right)$.
    
    \item Training of $\pitheta$ generally involves randomly sampling $\z \in \sZ$ and a transition $s_t,a_t,s_{t+1} \sim \mathcal{D}_{\textbf{train}}$. To perform an RL update with a conventional algorithm, we need to ``simulate'' the reward that would arise for state $s_{t+1}$ and for the reward function that the sampled $\z$ encodes. For this we need a decoding function: if $\epsilon(\{s_i, r(s_i)\}_{i=1}^ D) = \z_r$ then $\epsilon^{-1} (\z_r, s) \approx r(s)$. Since the space of reward functions contains more information than $\sR^d$, this is necessarily an approximation.
    Training unfolds by sampling a large amount of transitions and task vectors and executing RL updates of $\boldtheta$ with any RL algorithm.
\end{itemize}

\subsection{Bandit interactions: general objective}
The most generic RL objective would be the following: 
\begin{align*}
    \arg\max_{\Lambda}\mathbb{E}_{\text{interactions}}\left[\sum_{t=0}^{T-1}\gamma ^t R_{t+1}\right]
\end{align*} 
with ``interactions'' the interplay of probability distributions detailed in Section \ref{subsec:bandit}. 
However, this formulation would allow to pick state-dependent $Z_t$ to surpass the performance of $\pi$, for example by oscillating between vectors $\z_1, \z_2$ to stay at a state that the policy would not be able to stay at with a fixed $\z$. This would require the training of a task-specific agent, which would be much heavier than our bandit formulation.

\subsection{Linear Bandits} \label{app_subsec:linear_bandits}
Linear Bandits \citep{lattimore2020bandit} is an extension of a multi-armed bandit problem where an agent must quickly find the action yielding the highest reward. 
Critically, compared to RL, the optimal behavior is to choose a single action -- there is no temporal connection between steps, the environment only provides rewards. The crux of the problem lies in the \textit{exploration-exploitation dilemma} of picking the action we think is best versus an action we have little knowledge about.
Instead of a discrete amount of actions, in Linear Bandits the ``actions'' come as vectors $Z_t \in \sZ_t\subset \sR ^d $ and the reward is assumed to come as $R_t = Z_t \cdot \z_r + \eta_t $, with unknown $\z_r$ and generally $\eta_t \sim \N \left(0, \sigma^2\right)$. 
In this paper we study a specific version of the problem where $\sZ_t = \sZ$ is the sphere of radius $\sqrt{d}$ (actions are normalized) and $\z_r \in \sZ$. Under these assumptions, the optimal action is $Z_t=\z_r$.
The objective of Linear Bandits is to find an algorithm $\Lambda \left(Z_t \mid S_0 , Z_0 , R_1, \dots Z_{t-1}, S_t\right)$ that quickly maximizes reward, which is generally formalized as minimizing \textit{regret}: $\text{Reg}_T \doteq \sum _{t=1}^T \z_r\cdot \z_r - Z_t\cdot \z_r$, evaluating for each time-step how sub-optimal the choice of $Z_t$ was.

\subsection{Parallels with Linear Bandits}
\label{app_sec:bandit_comparison}
The regret definition for our setup is:
\begin{align}
	\text{Reg}_{T} = \sum_{t=0}^{T-1}\boldpsi\left(S_{t},\z_{r}\right)\cdot \z_{r}-\boldpsi\left(S_{t},Z_{t}\right)\cdot \z_{r}.
\end{align}
This Equation is reminiscent of Linear Bandits (LB, see Section \ref{app_subsec:linear_bandits}), in this Section we take a step back to compare our framing and objective to that of LB. 

\begin{itemize}
    \item In LB, the agent makes a decision and gets reward\footnote{We use different notations for concepts that are not directly comparable between LB and our setup} $X_t = D_t \cdot \z_r$, which it needs to maximize despite unknown $\z_r$. $D_t$ is both the action and the element we compare to $\z_r$ through a dot product to evaluate the action. 
    
    \item In our setup and Equation \eqref{eq:regret} however, our decisions, observations and rewards are decoupled:
    \begin{itemize}
        \item The ``action'' $Z_t$ induces $\boldpsi(S_t, Z_t)$ but is not equal or linearly linked to it, unlike $D_t$. It is $\boldpsi(S_t, Z_t)$ which is pitched against $\z_r$ to evaluate the quality of action $Z_t$, not $Z_t$ directly.
        \item $X_t$ is decoupled into the instantaneous reward $R_{t+1}=\boldphi(S_{t+1})\cdot \z_r + \eta_t$ and the value function in the regret, $\boldpsi (S_t,Z_t)\cdot \z_r$. After action $Z_t$ we observe the state-reward pair $(S_{t+1}, R_{t+1})$, but we are not trying to maximize instantaneous rewards. Note that observing $S_{t+1}$ allows us to observe $\boldphi(S_{t+1})$ and $\boldpsi(S_{t+1}, \z)$ for any $\z$.
        \item The least squares estimates $\hat{Z}_t$ are obtained from $\boldphi(S_t)$ but we aim to maximize the value function $\boldpsi(Z_t)\cdot \z_r$ 
    \end{itemize}
\end{itemize}

\section{Theoretical Results}
\subsection{Regret bound on USF-UCB} \label{app_sec:lemma_proof}
We aim to bind the total regret of USF-UCB as Algorithm \eqref{eq:USF-UCB}:
\begin{align*}
    \text{Reg}_T \leqslant 2\sum _{t=1} ^{T-1} \sqrt{\beta _t} \left\Vert \boldpsi \left(S_t, Z_t\right) \right\Vert_{V_{t} ^{-1}}.
\end{align*}

Consider the instantaneous regret $\text{reg}_t\doteq \boldpsi\left(S_{t};\z_{r}\right)\cdot \z_{r}-\boldpsi\left(S_{t};Z_{t}\right)\cdot \z_{r}$. Assuming, with high probability, that $\z_r \in \varepsilon_t$:
\begin{align*}
	 \text{reg}_t &=\boldpsi\left(S_{t};\z_{r}\right)\cdot \z_{r}-\boldpsi\left(S_{t};Z_{t}\right)\cdot \z_{r} \\
     & + \boldpsi\left(S_{t};\z_{r}\right)\cdot \hat{Z}_t - \boldpsi\left(S_{t};\z_r\right)\cdot \hat{Z}_t\\
     & = \boldpsi\left(S_{t};\z_{r}\right)\cdot \left( \z_{r}-\hat{Z}_t \right ) +\boldpsi\left(S_{t};\z_r\right)\cdot \hat{Z}_t - \boldpsi\left(S_{t};Z_{t}\right)\cdot \z_{r} \\
    \text{(CS)} &\leqslant \left\Vert \boldpsi\left(S_{t};\z_r\right)\right\Vert_{V_{t} ^{-1}}  \left\Vert \z_r - \hat{Z}_t \right\Vert_{V_{t}} +\boldpsi\left(S_{t};\z_r\right)\cdot \hat{Z}_t - \boldpsi\left(S_{t};Z_{t}\right)\cdot \z_{r} \\
    \text{(}\z_r \in \varepsilon_t\text{)} &\leqslant \sqrt{\beta _t} \left\Vert \boldpsi\left(S_{t};\z_r\right)\right\Vert_{V_{t} ^{-1}} +\boldpsi\left(S_{t};\z_r\right)\cdot \hat{Z}_t - \boldpsi\left(S_{t};Z_{t}\right)\cdot \z_{r} \\
    \text{(Equation \eqref{eq:USF-UCB})} &\leqslant \sqrt{\beta _t} \left\Vert \boldpsi\left(S_{t};Z_t\right)\right\Vert_{V_{t} ^{-1}} +\boldpsi\left(S_{t};Z_t\right)\cdot \hat{Z}_t - \boldpsi\left(S_{t};Z_{t}\right)\cdot \z_{r} \\
    \text{(same arguments)} &\leqslant2\sqrt{\beta_t}\left\Vert \boldpsi\left(S_{t};Z_{t}\right)\right\Vert _{V_{t}^{-1}}.
\end{align*}

\subsection{Preliminary work for the Elliptical Potential Lemma for USF} \label{app:elliptical_potential_lemma}
In Linear Bandits the regret is bounded by a sum of actions, $\sum _t\left\Vert A_t\right\Vert _{V_{t}^{-1}}$, which the Elliptical Potential Lemma bounds by the increase in determinant $\log \frac{\det V_T}{\det V_0}$, using the fact that $V_{t} \doteq V_{t-1} + A_t A_t^\top $. 
In our case because of the decoupling of actions $Z_t$, instantaneous observations $\boldphi(S_t)$ and the quantity we maximize, $\boldpsi(S_t,\z)\cdot \z_r$, finding an equivalent to the lemma is not trivial.

We can upper-bound the instantaneous regret naively, 
\begin{align*}
    \text{reg}_t &= \boldpsi\left(S_{t};\z_{r}\right)\cdot \z_{r}-\boldpsi\left(S_{t};Z_{t}\right)\cdot \z_{r} \\
    \text{(CS)} & \leqslant \left\Vert \boldpsi\left(S_{t};\z_{r}\right) - \boldpsi\left(S_{t};Z_t\right) \right\Vert _2 \left\Vert \z_r \right\Vert _2.
\end{align*}
We chose $\sZ$ as the sphere of radius $\sqrt{d}$ so $\left\Vert \z_r \right\Vert _2=\sqrt{d}$. We can similarly assume a maximum value on $\boldphi$, $\forall s \left\Vert \boldphi(s) \right\Vert _2 \leqslant \sqrt{d}\, \phi_{\max}$. From the definition of Successor Features as a discounted sum of features, this leads to 
\begin{align*}
    \text{reg}_t & \leqslant \frac{2d\phi _{\max}}{1-\gamma}.
\end{align*}
However this is not a very tight bound compared to exploiting the properties of $V$. Still, this can be combined with the previous bound to obtain 
\begin{align*}
    \text{reg}_t &\leqslant 2 \min \left(\sqrt{\beta _t}\left\Vert \boldpsi\left(S_{t};Z_{t}\right)\right\Vert _{V_{t}^{-1}}, \frac{d\phi _{\max}}{1-\gamma} \right).
\end{align*}

Such a minimum is often included in bandits to introduce a $\log$ operator. In our case the crux of the matter comes from the fact that we have a sum of norms of $\boldpsi$ instead of $\boldphi$, which prevents us from using the definition of $V$.

\subsection{Maximization of the elliptical norm} \label{app:more_ellipsis}
Our USF-UCB algorithm (Equation \eqref{eq:USF-UCB}) proposes an exploration component through the maximization of the elliptical norm $\left\Vert \boldpsi\left(S_{t};Z_{t}\right)\right\Vert _{V_{t}^{-1}}$. We show now that several other approaches could make this quantity arise, highlighting its key importance. 

\paragraph{Maximization on the Confidence Bound.}
From the confidence interval $\varepsilon_t$ in Equation \eqref{eq:ellipsis}, a more direct algorithm would be to apply optimism in the face of uncertainty on the expected return:
\begin{align}
	Z_t &= {\arg \max} _{\z \in \varepsilon_t} \boldpsi\left(S_{t}; \z\right)\cdot \z. \label{eq:UCB1}
\end{align}

The regret bound is the same as our method: assume, with high probability, that $\z_r \in \varepsilon_t$, then,
\begin{align}
	\rho_{t}	&\doteq \boldpsi\left(S_{t};\z_{r}\right)\cdot \z_{r}-\boldpsi\left(S_{t};Z_{t}\right)\cdot \z_{r} \\
	&\leqslant\boldpsi\left(S_{t};Z_{t}\right)\cdot Z_{t}-\boldpsi\left(S_{t};Z_{t}\right)\cdot \z_{r} \\
	&=\boldpsi\left(S_{t};Z_{t}\right)\cdot\left(Z_{t}-\z_{r}\right) \\
	&\leqslant\left\Vert \boldpsi\left(S_{t};Z_{t}\right)\right\Vert _{V_{t}^{-1}}\left\Vert Z_{t}-\z_{r}\right\Vert _{V_{t}}\\
	&\leqslant2\left\Vert \boldpsi\left(S_{t};Z_{t}\right)\right\Vert _{V_{t}^{-1}}\beta_t. \label{eq:regret_bound_ucb}
\end{align}
Unlike our algorithm, in this case we need $Z_t \in \varepsilon_t$ because it is used to evaluate the return, not only as an action to input in $\boldpsi$. It is sometimes clearer to rewrite $Z_t = \hat{Z}_t + \beta _t \boldo _t$, with $\hat{Z}_t$ the ``greedy'' part and $\boldo _t$ the ``optimistic'' part of the vector, constrained to  $\left\Vert \boldo_{t}\right\Vert _{V_{t}} \leqslant 1$:

\begin{align*}
    Z_t&= \underset{\substack{
\z=\hat{Z}_t+\sqrt{\beta_t}\boldo_t\\
\left\Vert \boldo_t \right\Vert _{V_{t}} \leqslant 1
}}{\arg \max} \boldpsi\left(S_{t}; \z\right)\cdot \hat{Z}_t + \sqrt{\beta _t} \boldpsi\left(S_{t}; \z\right)\cdot \boldo_t.
\end{align*}

While this approach has similarities with ours, it is more cumbersome and less clear.

\paragraph{Variance Minimization}
Through our USF framework we have that $R_{t} = \boldphi (S_t) \cdot \z_r + \eta_t$, with $\eta_t \sim \mathcal{N} (0,\sigma^2)$. Writing $\Phi$ the matrix of columns $\phi(S_t)$ and $\boldr$ the vector of rewards $R_t$, the least squares estimator $\hat{Z}_t = V_t ^{-1} \Phi ^\top \boldr$ therefore has covariance \begin{align*}
\text{Cov} (\hat{Z}_t) &= \E \left[ (\hat{Z}_t - \z_r) (\hat{Z}_t - \z_r)^\top \right] \\
& = V_t ^{-1} \Phi ^\top \sigma ^2 \Phi V_t ^{-1} \\
& =\sigma^2 V_{t}^{-1}
\end{align*} 
and the value function estimator $\boldpsi (S_t,\z) \cdot \hat{Z}_t$ has variance 
\begin{align*}
    \text{Var} (\boldpsi (S_t, \z) \cdot \hat{Z}_t) &= \boldpsi (S_t, \z)^\top \text{Cov} (\hat{Z}_t) \boldpsi (S_t, \z)\\
    &= \sigma^2 \boldpsi (S_t, \z)^\top V_{t}^{-1} \boldpsi (S_t, \z)
\end{align*}
We find again the (squared) elliptical norm, which geometrically indicates the length of the Successor Features weighted by the strength of uncertainty in each feature direction. Maximizing this norm over $\z$ means picking the policy that will spend the most time in unexplored regions of the feature space.

\paragraph{Optimal Experiment Design}
We shift the focus away from regret minimization and directly focus on finding the $\z_r$ which will reduce uncertainty most; from an RL/bandit perspective, this is tantamount to focusing on pure exploration. There are several metrics for uncertainty, virtually always involving the eigenvalues of $V_t$ or its inverse. One of particular interest for us is D-optimal design, where we look to maximize the determinant (product of eigenvalues) of $V_t$: $\max _{\Lambda} \log \det V_t$. Considering that $V_{t+1}= V_t + \boldphi (S_{t+1}) \boldphi (S_{t+1})^\top$, a common approach is to use the matrix determinant lemma:
\begin{align*}
    \det (V_{t+1}) = \det (V_t)(1+\boldphi (S_{t+1})^\top V_t ^{-1} \boldphi (S_{t+1})),
\end{align*}
so maximizing the step-wise increase in determinant is equivalent to maximizing $\boldphi (S_{t+1})^\top V_t ^{-1}\boldphi (S_{t+1})$. We find again the elliptical norm, this time for a single time-step, measuring how informative a single state is. Using this term directly as an intrinsic reward is a common approach in linear bandits and has been applied to RL for exploration before. 

The critical insight from using USF is that instead of giving an intrinsic bonus to $S_{t+1}$, we want to give an intrinsic bonus to an entire \textit{policy} encoded by $\z$. So we are looking for the policy which generates the most of this one-step intrinsic bonus. At step $S_t$ with $V_{t}$, we can think about the \textit{future certainty matrix}, the features that will be visited by policy $\pi (\cdot\mid  S_t; \z)$ during its trajectory: $M_t (\z) = \E _{\pi (\cdot; \z)} \left[ \sum _{k=1}^{\infty} \gamma ^{t+k} \boldphi(S_{t+k+1}) \boldphi(S_{t+k+1})^\top \right]$. We are interested in the matrix $V_t + M_t(\z)$, which encapsulates the whole trajectory (before and after $t$ respectively), and in particular in maximizing its determinant. 

Though we do not have access to $M_t$ directly, $\boldpsi$ encodes related information -- as a reminder, $\boldpsi (S_t, A_t; \z) = \E _{\pi (\cdot; \z)} \left[ \sum _{k=1}^{\infty} \gamma ^{t+k} \boldphi(S_{t+k+1}) \right]$. Therefore the proposal is to use $\boldpsi(S_t; Z_t) \boldpsi(S_t; Z_t)^\top$ as a rank-1 approximation of $M_t$, with the intuition that $\boldpsi$ summarizes future feature visits. Using the matrix determinant lemma again gives 
\begin{align*}
    \det (V_{t}+\boldpsi (S_t; Z_t) \boldpsi (S_t; Z_t)^\top) &= \det (V_t) \left(1 + \boldpsi (S_{t}, Z_t)^\top V_t^{-1} \boldpsi (S_{t},Z_t)\right)
\end{align*}
so under this approximation we need to maximize $\boldpsi (S_{t}, Z_t)^\top V_t^{-1} \boldpsi (S_{t},Z_t)$ to maximally gain information into $V_t$.

The issue with this link is in the approximation - it is not exactly aligned with truly predicting future features, as the summation averages out directions in feature space (mismatch between $M_t$ and $\boldpsi \boldpsi ^\top$). This perspective potentially motivates learning $M_t$ in addition to $\boldpsi$ in the USF framework, which would allow access to such covariance-based estimators at transfer time.


\section{Experimental details}
\subsection{Example choices for SF: Successor Clusters and Laplacian Eigenfunctions}
\label{app_sec:lap_clusters}
We now detail the two types of features which we experiment with in this paper, but our results in Section \ref{sec:method} apply to any choice of $\boldphi$. We visualize the features proposed below, and some of the corresponding policies from a trained USF Approximator, in Figure \ref{fig:explo}.

\paragraph{Successor Clusters} \citep{bagot2025successorclusters} propose an intuitive choice of $\boldphi$: since we need to linearly approximate the reward function, one possibility is to discretize it over a partition of the state space. This is achieved by breaking the state space down into clusters $\left\{C_i\right\}_{i=1}^d$ with a temporal notion of distance, and pick \textit{cluster occupancy} features: $\phi_i(s) = \boldsymbol{1}_{s\in C_i}$. The resulting encoding procedure leads to $z_i$ containing the average reward in cluster $C_i$.
This choice of features also implies that the associated Successor Features define a specific Successor Measure \citep{blier2021learning}: $\psi_{\pi} (s,a)_i$ indicate the expected time spent in cluster $C_i$ when following policy $\pi$. 
This property is very convenient for interpretability, and in general Successor Clusters allow us to understand and visualize several key elements of the SF/USFA framework. For example, as a critical intuition, for generating exploration policies we can simply count the number of visits to each cluster, $N(C_i)$, and set $z_i = 1/N(C_i)$. This borrows from MBIE-EB \citep{strehl08mbieeb} with the idea that clusters visited more often get less of a bonus.

\paragraph{The Laplacian Eigenfunctions} have long been argued to play a key role in Reinforcement Learning \citep{mahadevan2007proto, machado2017laplacian}, in particular in recent years for generating intrinsic rewards, options and zero-shot policies with Successor Features \citep{barreto2019option,machado2021temporal,touati2023does}. The \textit{graph Laplacian} is defined as the matrix $L=D-A$, with $D$ the degree matrix and $A$ the adjacency matrix of the MDP under some policy $\pi$, usually random uniform. The ``Laplacian Representation'' in RL refers to the first $d$ eigenvalues of the (often normalized) Laplacian, $\boldphi (s) = \left[f_1(s), \dots, f_d(s) \right]$, with $f_i(s)$ the ith eigenfunction evaluated at state $s$. The notation with $\boldphi$ is intentional: we can plug this directly into the SF/USFA framework. The appeal for the Laplacian Representation comes from the fact that it behaves like a Fourier-like basis over reward functions, with functions of increasing frequency along the smooth temporal directions of the state space. Several methods exist to compute approximate eigenfunctions in the context of complex state and action spaces \citep{wu2018laplacian,wang2021towards,gomez2024proper}.

\subsection{Hyper-parameters and experimental choices}
\textbf{$\boldphi$ features.} We discussed the clusters and Laplacian features above, in practice we simply cut the $9\times 9$ state space into $9$ clusters of size $3 \times 3$ for the cluster features, for the Laplacian eigenvectors we directly create the matrix and compute its eigenvectors using standard libraries. 

\textbf{USFA architecture and training.} \label{app_sec:usfa}
The USFA $\boldpsi_{\boldtheta} (s,a;\z)$ is a fully connected neural network with 4 hidden layers of size $256$. We find a dropout rate of $0.15$ to help with generalization across $\z$. 
We train the USFA to predict the optimal sum of future $\boldphi (S_{k\geqslant t})$ to maximize expected sum of rewards  $R_k=\boldphi(S_k)\cdot \z$, by repeatedly sampling $\z$ from the $\sqrt{d}$-diameter ball and minimizing the Bellman gap with a MSE loss. We use soft target updates with $\tau=0.01$ and the Adam optimizer with a learning rate of $0.00025$. We choose $\gamma=0.99$ for accurate future feature predictions. The USFA weights are frozen before the start of the online transfer.

\subsection{``Exhaustive'' Explorer}
\label{app:exhaustive}
\begin{figure}
    \centering
    \includegraphics[width=0.25\linewidth]{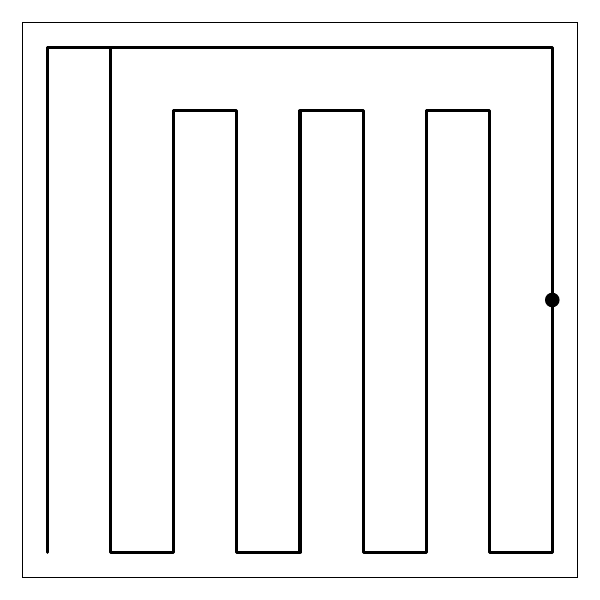}
    \caption{Example trajectory of the exhaustive explorer, covering the domain by visiting all states.}
    \label{fig:exhaustive}
\end{figure}

To compare our methods we implement an exploring agent that reaches all states. In a simple gridworld this is easy to do: we simply count state visits and use a greedy one-step look-ahead to find which action visits the least-visited state next. In Figure \ref{fig:exhaustive} we show an exploration trajectory generated by this method. 

\end{document}

%% file: my_commands.tex
\newcommand{\boldphi}{\boldsymbol{\phi}}
\newcommand{\boldpsi}{\boldsymbol{\psi}}
\newcommand{\boldtheta}{\boldsymbol{\theta}}
\newcommand{\pitheta}{\pi_{\boldtheta}}

\newcommand{\boldr}{\boldsymbol{r}}

\newcommand{\z}{\boldsymbol{z}}

\newcommand{\boldo}{\boldsymbol{o}}

\def\sS{{\mathcal{S}}}
\def\sA{{\mathcal{A}}}
\def\sO{{\mathcal{O}}}
\def\sZ{{\mathcal{Z}}}
\def\sR{{\mathbb{R}}}
\def\gM{{\mathcal{M}}}

\def\E{{\mathbb{E}}}
\def\N{{\mathcal{N}}}

\usepackage[dvipsnames]{xcolor}

%% file: fig_pretrain.tex
\begin{figure}[p]
	\centering
	\begin{subfigure}{\textwidth}
		\centering
		\includegraphics[width=\textwidth]{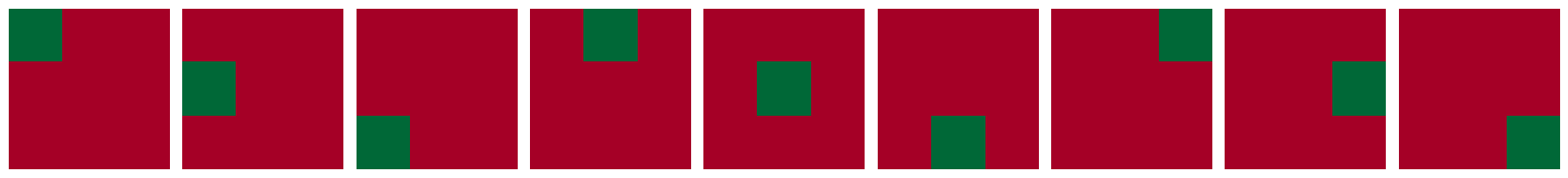} \\ 
		\includegraphics[width=\textwidth]{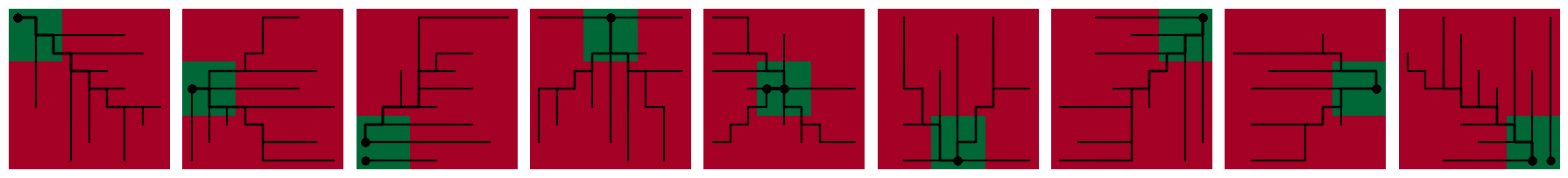} \\ 
		\includegraphics[width=\textwidth]{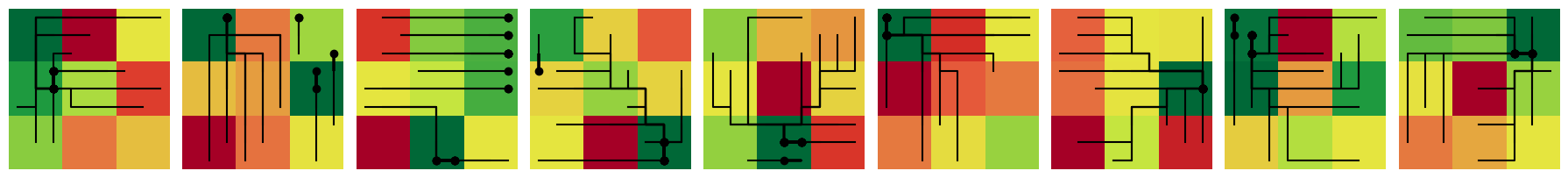}
		\caption{Cluster occupancy features and associated USFA policies.}
		\label{fig:features_usfa_clusters}
	\end{subfigure}
	\begin{subfigure}{\textwidth}
		\centering
		\includegraphics[width=\textwidth]{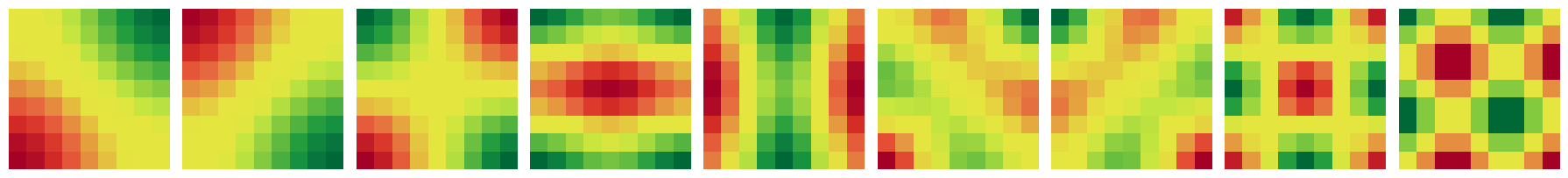}
		\includegraphics[width=\textwidth]{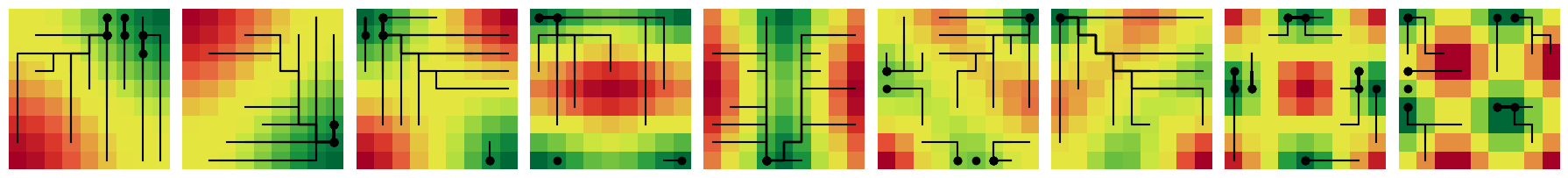}
		\includegraphics[width=\textwidth]{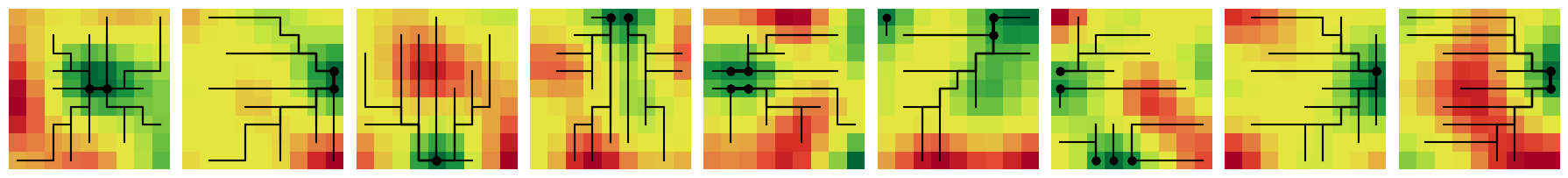}
		\caption{Laplacian eigenfunction as features and associated USFA policies.}
		\label{fig:features_usfa_lap}
	\end{subfigure}
	\caption{Visualization of the pre-training setup on top of which we want to perform online transfer. The environment is a simple $9\times 9$ gridworld with actions \texttt{up}, \texttt{down}, \texttt{left}, \texttt{right}. In the Universal Successor Features framework the agent (the USF Approximator, USFA) learns to produce optimal policies for all rewards expressed as $r(s)=\boldphi(s)\cdot \z_r$, with $\boldphi$ a fixed basis over reward functions (``features'') and $\z_r \in \sZ \subset \sR^d$ any weight vector. \\
    We visualize our two choices of features $\boldphi$ at the top of each Figure (Figure (a): cluster occupancy, Figure (b): Laplacian eigenfunctions). In the middle row we visualize some associated trajectories generated by the USFA for the one-hot vectors $\z=\boldsymbol{e_i}=[0,\dots,0, 1,0,\dots,0]$, while the bottom row showcases randomly sampled $\z \sim \sZ$. The reward functions as $\phi \cdot \z$ are shown as background colors over the 2D state space ranging from red (minimum) to green (maximum), and the agent is expected to reach green regions while avoiding red ones. The trajectories generated by the USFA are shown as lines in black, starting from a random state and ending at a point marked as a solid circle.
	}
	\label{fig:features_usfa}
\end{figure}

\begin{figure}[p]
    \centering
    \includegraphics[width=\linewidth]{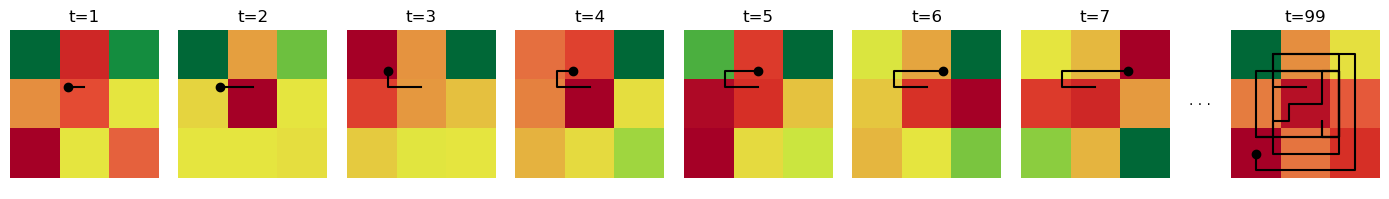}
    \includegraphics[width=\linewidth]{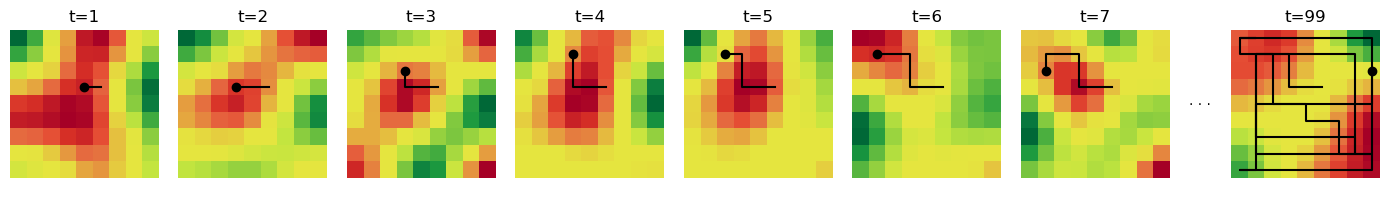}
    \caption{Samples of pure exploration patterns for our USF-UCB algorithm on our two feature choices (top: clusters, bottom: Laplacian). At each time-step, USF-UCB (Equation \eqref{eq:USF-UCB} with $\hat{Z}_t = \boldsymbol{0}$) recommends the next $Z_t$ to try, for which we show the associated reward function ($\boldphi \cdot Z_t$) in the background. $Z_t$ is passed to the BFM to produce the next action, repeatedly until $T=99$ steps (rightmost square). In both trajectories, the $Z_t$ incentivizes exploration by rewarding regions that have not been visited.}
    \label{fig:explo}
\end{figure}

%% file: fig_expes.tex
\newlength{\figsize}
\setlength{\figsize}{.43\textwidth}
\begin{figure}
    \centering
    \includegraphics[width=\figsize]{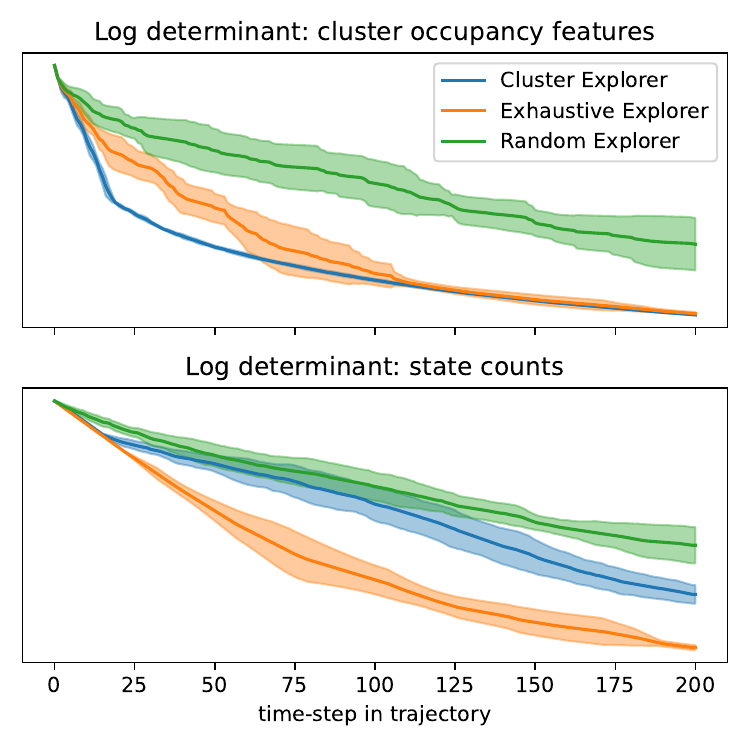}
    \hfill
    \includegraphics[width=\figsize]{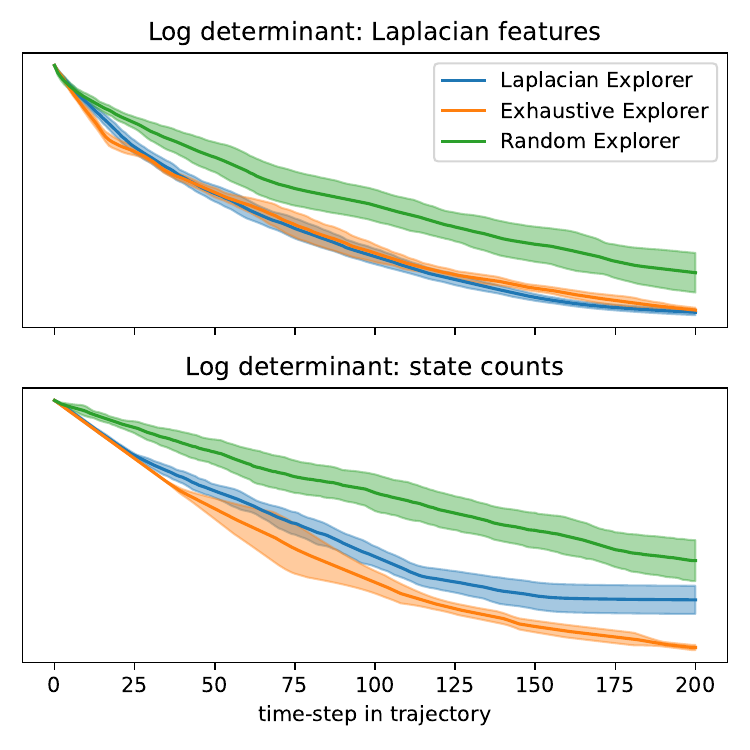}
    \caption{Efficiency of exploration during a trajectory as measured by the log determinant of the uncertainty matrix, computed either with respect to the features (top row) or the state counts (bottom row). We pit our method against an ``exhaustive'' explorer (visits all states) and a random explorer for both feature choices (left: clusters, right: Laplacian). Our explorers are comparable or better than the exhaustive explorer with respect to their own feature space, but not at reaching all states.}
    \label{fig:determinants}
\end{figure}

\begin{figure}
    \centering
    \includegraphics[height=0.08\textheight]{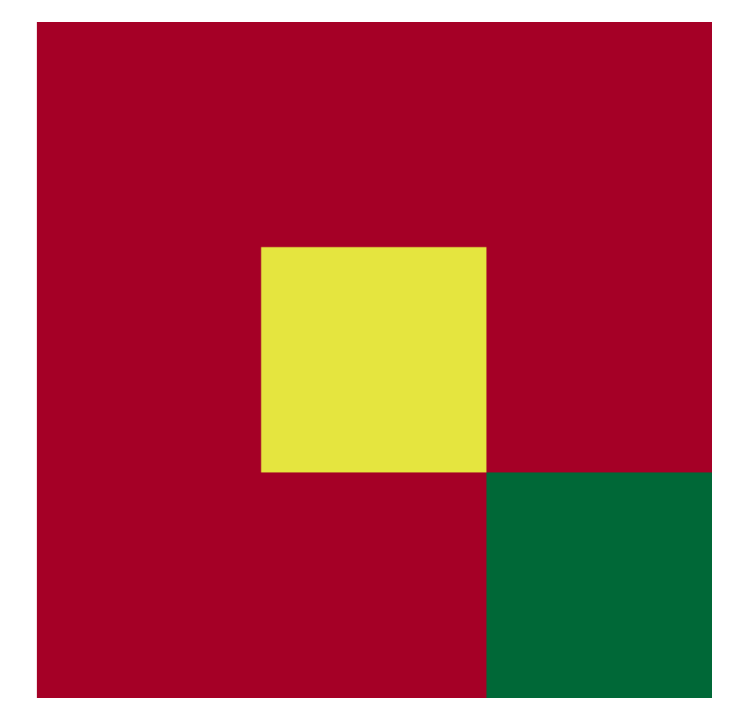}
    \vrule
    \includegraphics[height=0.08\textheight]{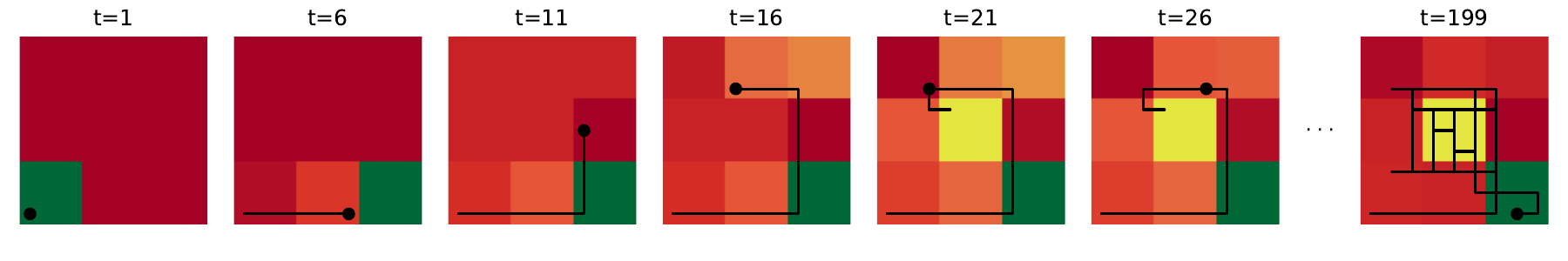}
    \includegraphics[height=0.08\textheight]{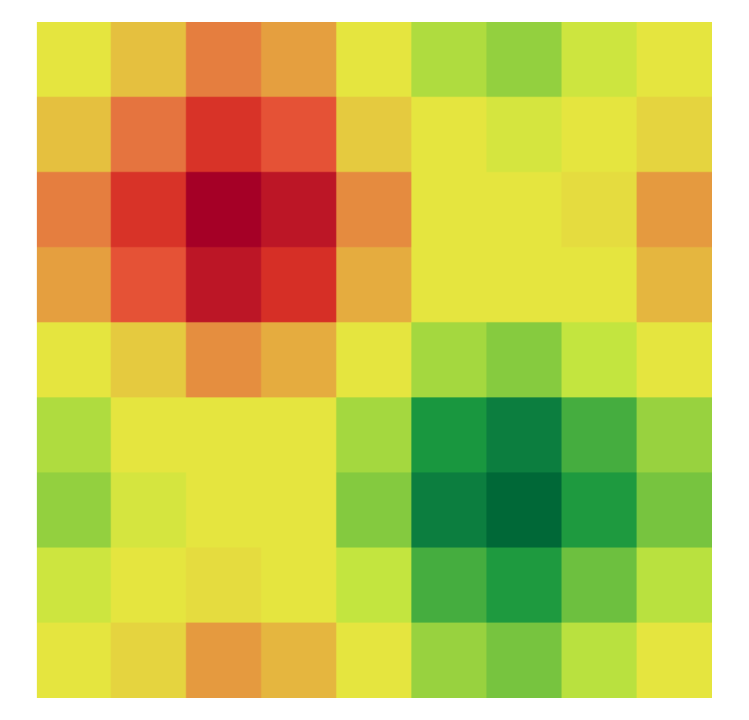}
    \vrule
    \includegraphics[height=0.08\textheight]{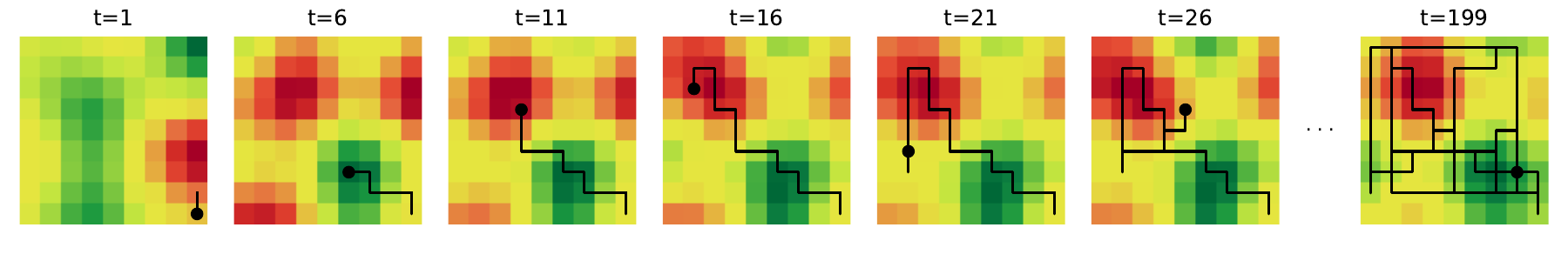}
    \caption{Visualization of online transfer trajectories. On the left, we display the reward function associated with the task vector $\z_r$ to find. The agent needs to interact with the domain to maximize this reward. We then display the trajectory, every $5$ steps, as the agent explores and eventually exploits. In the background we now show only the linear least squares estimate $\hat{Z}_t$. We can see the agent's estimation improve despite the added noise; the next Figure quantifies this.}
    \label{fig:samples_exploit}
\end{figure}

\begin{figure}
    \centering
    \includegraphics[width=\figsize]{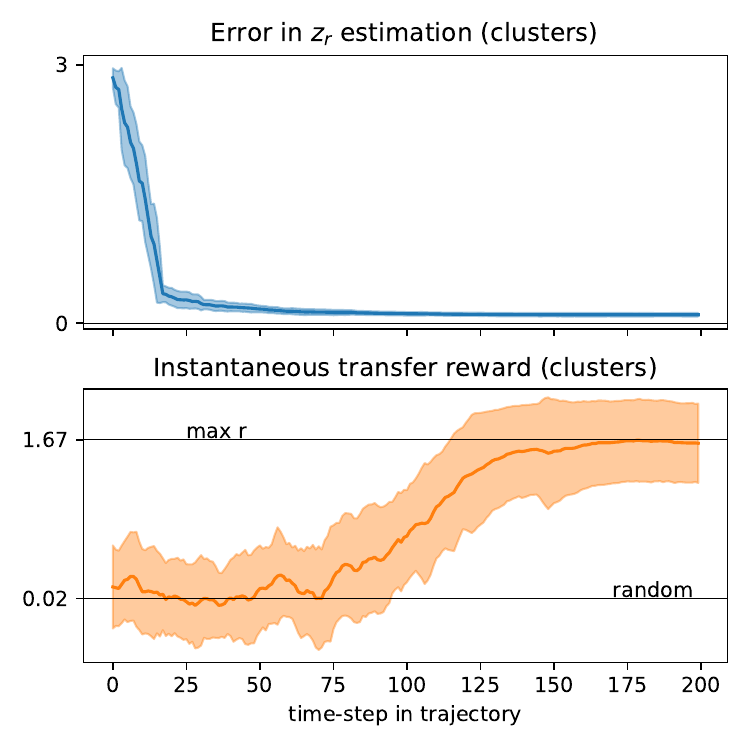}
    \hfill
    \includegraphics[width=\figsize]{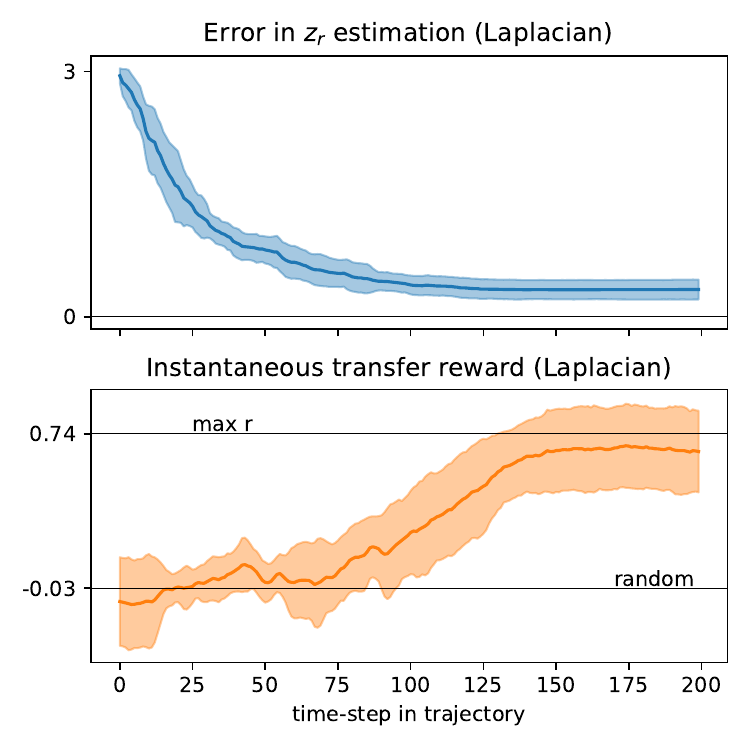}
    \caption{Evaluation of online transfer: (top) error in the estimation of $\z_r$ through $\hat{Z}_t$ (L2 distance), (bottom) instantaneous reward throughout the trajectory, for both feature choices (left: clusters, right: Laplacian). The estimator converges and the method finds the highest-rewarding states (\texttt{max r}). 
    }
    \label{fig:exploit}
\end{figure}